\documentclass{article}
\usepackage{amsmath}
\usepackage{graphicx}
\usepackage{amsfonts}
\usepackage{amssymb}
\begin{document}

\title{Using Bayesian Blocks to Partition Self-Organizing Maps}
\author{
Paul R. Gazis\footnote{SETI Institute, NASA Ames Research Center, 
Moffett Field, California, 94035, USA}
and Jeffrey D. Scargle\footnote{NASA Ames Research Center, Moffett Field, 
California, 94035, 
USA}}
\maketitle

{\em Abstract.}
Self organizing maps (SOMs) are widely-used for unsupervised classification.
For this application, they must be combined with some partitioning scheme 
that can identify boundaries between distinct regions in the maps they 
produce.  We discuss a novel partitioning scheme for SOMs based on the
{\em Bayesian Blocks} segmentation algorithm of {\em Scargle} [1998].  This 
algorithm minimizes a cost function to identify contiguous regions over 
which the values of the attributes can be represented as approximately 
constant.  Because this cost function is well-defined and largely 
independent of assumptions regarding the number and structure of clusters in 
the original sample space, this partitioning scheme offers significant 
advantages over many conventional methods.  Sample code is available.

\section{Introduction}

Self organizing maps (SOMs) 
[{\em Kohonen}, 1984, 1997; {\em Ritter et al.}, 1992] 
have seen considerable use as a means of unsupervised classification.  The 
SOM algorithm maps samples in some N-dimensional sample space, 
${\cal R}^{N}$, into an array of cells or principal elements (PEs) in a 
classification space, $\cal{A}$, of reduced dimensionality (usually one or 
two dimensions) in a way that reproduces, as much as possible, the 
topological structure of the original distribution.  In particular, the SOM 
algorithm will attempt to map adjacent clusters in the original sample space 
into adjacent blocks of contiguous PEs.  A variety of measures have been 
proposed to evaluate the degree to which topology is preserved by a 
particular mapping 
[{\em Villmann et al.}, 1997; {\em Bauer et al.}, 1999; 
{\em Hsu and Halgamuge}, 2001].

Used alone, SOMs serve as a means to visualize complicated relationships 
between groups of samples.  For classification purposes, they must be 
combined with some partitioning scheme that can identify boundaries of 
regions in the map that correspond to different clusters in the original 
sample space.  Early approaches involved comparing changes between adjacent 
PEs with some predetermined threshold
[{\em Ultsch and Simmons}, 1992 and {\em Merkl and Rauber}, 1997].
Unfortunately, it is not always easy to determine the proper threshold from 
actual data.  This can limit the utility of threshold-based schemes,
particularly in the case of poorly-defined or gradual boundaries.  More
sophisticated approaches can be divided into 'partitive' schemes, which 
seek to minimize some error function based on assumptions regarding the 
number and shape of clusters in the sample space, and 'hierarchical' 
schemes, based on analysis of the 'dendrogram' or tree structure of 
distances between adjacent PEs.
{\em Vesanto and Alhoniemi} [2000] and
{\em Ta\c{s}demir and Mer\'{e}nyi} [2005]
provide discussion and a comparative review of some of these approaches.  
Other investigators have proposed modifications to the SOM algorithm
itself 
[{\em Bauer and Villman}, 1996; {\em Alahakoon et al.}, 2000].
In recent years, hierarchical schemes have been gaining favor.  While these 
may be less ad hoc than many alternatives, they can still be vulnerable to 
local peculiarities of the data due to statistical fluctuations or other 
causes that alter the structure of the dendrogram, particularly near the
root nodes.

The {\em Bayesian Blocks} algorithm
[{\em Scargle}, 1998, 2001; {\em Scargle et al.}, 2003] 
is a segmentation algorithm based on the statistics of the data set that is
largely independent of assumptions regarding the nature of clusters in the 
classification space.  For these reasons, it offers advantages over schemes 
that involve the use of ad hoc user-defined parameters.  In Bayesian Blocks,
the space is partitioned into contiguous segments or 'blocks' and the values 
of the attributes are taken to be constant within each segment.  In the case 
of a SOM, the space to be partitioned will be the map itself and the 'blocks' 
will consist of clusters of contiguous PEs.  (Note that this is subtly 
different from the 'conventional' Bayesian Blocks approach, in which 
partitioning is performed in the original sample space.)

The following discussion describes a Bayesian Blocks-based partitioning 
scheme for SOMs.  Section 2 describes the mathematical basis of the cost
function.  Section 3 describes the implementation of this cost function in a 
partitioning algorithm.  Section 4 presents the results of tests on a sample 
data set to demonstrate the feasibility of this approach.  Section 5 
presents conclusions and suggestions for future work.

\subsection{A Note on Nomenclature}

Different communities use a wide variety of different terms to refer to 
clusters of data and the elements of which they are composed.  For reasons
of clarity and consistency, the following terms will be used throughout:

\begin{itemize}
\item{\bf PE:} The basic unit in the SOM output space. For consistency
with the extensive literature on SOMs, this is used in preference to
terms such as 'cell' or 'bin'.

\item{\bf block:} A collection of PEs, modeled as having a constant 
value of some quantity.  In general, these will belong to the same
cluster or class.

\item{\bf cost function:} The cost associated with partitioning a SOM
into a particular arrangement of blocks.  Synonymous with terms such as 
'fitness measure', 'risk', or 'objective function'.
\end{itemize}

\section{Mathematical Basis of the Cost Function}

The cost function is central to the Bayesian Blocks algorithm.  In the 
conventional Bayesian Blocks approach, this cost function is based on a 
model in which the density of samples in the sample space is taken as 
constant.  For a SOM, the situation is more complex, and the cost function 
must be based on the likelihood of a block of PEs in the classification 
space as a model for a cluster in the sample space.  The form of this cost 
function may not be intuitive, and experience has shown that it is all too 
easy to overlook or neglect terms during its formulation.  For this reason, 
the derivation will be discussed in some detail, beginning with the simple 
case of the cost function for two PEs, 1 and 2, with a single attribute, x.  
If these two PEs are associated with a single block, J, with value attribute 
value $x_J$, then assuming a normal distribution for the attributes, the 
probability that they will have values $x_{1}$ and $x_{2}$ will be:
\begin{eqnarray}
\lefteqn{P( x_{1}, x_{2} | J, x_{J}) dx_{1} dx_{2} = } \label{eq:P_xJ} \\
& & \frac{1}{\pi\sigma_{1}\sigma_{2}} 
\exp\left( -\frac{(x_{1} - x_{J})^2}{\sigma_{1}^2}\right) 
\exp\left( -\frac{(x_{2} - x_{J})^2}{\sigma_{2}^2}\right) 
dx_{1} dx_{2} \nonumber
\end{eqnarray}
where $x_i$ and $\sigma_i$ are the mean value and variance of the attribute 
in cell i.  One can expand this expression to separate the factors that 
vary with $x_{J}$ to obtain:
(\ref{eq:P_xJ}) to obtain:
\begin{eqnarray}
\lefteqn{P( x_{1}, x_{2} | J, x_{J}) dx_{1} dx_{2} = } \label{eq:P_X_xJ} 
\\
 & & \frac{1}{\pi\sigma_{1}\sigma_{2}}
\exp\left( -\frac{(x_{1} - X)^2}{\sigma_{1}^2}\right) 
\exp\left( -\frac{(x_{2} - X)^2}{\sigma_{2}^2}\right) \nonumber \\
 & & \exp\left( -\frac{(x_{J} - X)^2}{\sigma_{1}^2} -\frac{(x_{J} - 
X)^2}{\sigma_{2}^2}\right) dx_{1} dx_{2} \nonumber
\end{eqnarray}
where
\begin{displaymath}
X =
\left(\frac{x_1}{\sigma_1^2}+\frac{x_2}{\sigma_2^2}\right)
\left(\frac{1}{\sigma_1^2}+\frac{1}{\sigma_2^2}\right)^{-1}
\end{displaymath}
If the probability $P(x_J)dx_J$ is assumed to be uniform over a range R that
is large compared to $\sigma_i$, then one can integrate Equation 
(\ref{eq:P_xJ}) over $x_J$ to obtain:
\begin{eqnarray}
\lefteqn{P( x_{1}, x_{2} | J) dx_{1} dx_{2} \approx } \\
 & & \frac{1}{\pi\sigma_{1}\sigma_{2}}
\exp\left( -\frac{(x_{1} - X)^2}{\sigma_{1}^2}\right) 
\exp\left( -\frac{(x_{2} - X)^2}{\sigma_{2}^2}\right) \nonumber \\
 & & 
\frac{\sqrt{\pi}}{R}\left(\frac{1}{\sigma_1^2}+\frac{1}{\sigma_2^2}\right)^{-
1/2}
dx_{1} dx_{2} \nonumber
\end{eqnarray}
in which the quantity $x_J$ has been treated as a nuisance parameter and 
marginalized, as in the usual Bayesian formalism.  This expression can be 
generalized to N cells with a set $\{x_i\}$ of attribute values where 
$i\in\{1...N\}$
\begin{eqnarray}
\lefteqn{P( \{x_i\} | J) \approx } \label{eq:P_set_xi} \\
 & & \left(\prod_{i=1}^N \sigma_i \right)^{-1}
 \exp\left( -\sum_{i=1}^N\frac{(x_{i} - X)^2}{\sigma_{i}^2}\right) 
 \frac{1}{(\sqrt\pi R)^{N-1}}
\left(\sum_{i=1}^N\frac{1}{\sigma_i^2}\right)^{-1/2} \nonumber
\end{eqnarray}
where X is now
\begin{displaymath}
X =
\left(\sum_{i=1}^N\frac{x_i}{\sigma_i^2}\right)
\left(\sum_{i=1}^N\frac{1}{\sigma_i^2}\right)^{-1}
\end{displaymath}

Equation (\ref{eq:P_set_xi}) describes the conditional probability, 
$P( \{x_i\} | J)$, that the N cells in a region J will have the values
$\{x_i\}$.  One can apply Bayes Theorem to obtain the probability, 
$P(\{J_k\} | \{x_i\})$ that a set of $N_{blocks}$ blocks, $\{J_k\}$, can 
be created from a set of N cells with a set of attribute values, $\{x_i\}$.
\begin{equation}
P(\{J_k\} | \{x_i\}) = \frac{P(\{x_i\} | \{J_k\})P(\{J_k\})}{P(\{x_i\})}
\end{equation}

In the absence of any reason to prefer a particular distribution of blocks, 
the prior probabilities, $P(\{J_k\})$ will can be neglected.  The prior 
probabilities, $P(\{x_i\})$ can also be neglected, since one is only 
interested in the relative probability of different configurations of 
blocks.  This can be converted to a cost function, $C(\{x_i\},\{J_k\})$, for 
a partitioning ${J_k}$ by taking the negative logarithm of the product of 
the cost functions for each block:
\begin{eqnarray}
C(\{x_i\},\{J_k\}) & = & \sum_k \left[ ( N_k - 1) ( \ln( R) + \ln\sqrt\pi) - 
\label{eq:cost_1} \right. \\
& & - \left. \left( \ln\left(\prod_{i=1}^{N_k} \sigma_i \right) +
\ln\left(\sum_{i=1}^{N_k} \frac{1}{\sigma_i^2}\right) \right) - 
\left(\sum_{i=1}^{N_k} \frac{x_i^2}{\sigma_i^2} - X_k\right) \right] \nonumber
\end{eqnarray}
where $N_k$ and $X_k$ are the values for the individual blocks, $J_k$
of the partition.  Note that in this expression, the sums and products of
$\sigma_i$ for each block are taken over the $N_k$ PEs of the block itself
rather than the entire set of $N$ PEs as a whole.

This cost function has three terms.  The first term accounts for the effect 
of the range parameter, R.  Increasing this parameter will reduce the 
likelihood that similar values of the attribute might occur by chance, and 
increase the cost of joining a set of regions.  The second term accounts 
for uncertainties in the values of the attribute.  As this uncertainty
increases, the cost of joining a set of regions will go down.  The final 
term accounts for the effect of variations between regions.  The cost 
function in Equation (\ref{eq:cost_1}) can be generalized to multivariate 
data with M attributes, $\vec{x}$ in a straightforward fashion:
\begin{equation}
C(\{\vec{x_i}\},\{J_k\}) = \sum_{j=1}^{M} C(\{x_i^{(j)}\},\{J_k\}) 
\label{eq:total_cost}
\end{equation}

In principle, there are several ways that this expression can be used.
One can calculate cost functions for individual blocks and sum them to
compare the cost functions for different partitionings.  One can calculate
the cost function for a partitioning and compare it with the cost of a
contiguous region.  One can even combine statistics for all of the PEs in
each block to calculate values of the mean and standard deviation for that
block and treat each possible partitioning as a collection of blocks of
size one.  It can be shown that these different approaches are 
mathematically equivalent.

Two issues complicate determination of the cost function.  One involves 
the variances, $\vec{\sigma}_i$, of the attribute values within a PE.  In 
general, these will not be available, and the variation might not even 
follow a normal distribution.  In practice, it will be necessary to 
estimate $\vec{\sigma}_i$ from the set of observed standard deviations, 
$\vec{s_i}$, for the samples in that PE.  Because the SOM algorithm 
ensures that adjacent PEs will not overlap, one can represent them as 
unevenly-spaced bins.  If the widths, $\{\vec{\Delta x}_i\}$, of these bins 
are sufficiently small, the distribution within each PE will be uniform, 
and it is straightforward to show that $s_i = \Delta x_i / 12$.  We
arbitrarily assume that for the $i_{th}$ PE of block J, 
$\sigma_i \approx N_{scale} \Delta x_i$, where $N_{scale}$ is some 
appropriate scale size for that block, such as the square root of the 
number of PEs it contains.  Then 
$\vec{\sigma}_i \approx 12N_{scale} \vec{s}_i$.  
In practice, partitionings turn out to be comparatively insensitive to the
$\vec{\sigma}_i$, and it will be shown that these parameters can be 
allowed to vary by almost an order of magnitude with little effect on the 
results.

The second issue involves the range, R, over which the attribute values, 
$\vec{x}_J$, for the blocks themselves can vary.  As was the case for the 
variances within each block, this information may not be available, but it 
is possible to obtain an estimate from the observed mean values of the 
attributes for the associated PEs.  In practice, a suitable approximation is 
simply $R = 2 ( x_{max} - x_{min})$.  As with the standard deviation, 
partitionings are comparatively insensitive to the value of this parameter, 
but it should be noted that the cost function in Equation (\ref{eq:cost_1}) 
involved the assumption that the distribution, $P(x_J)dx_J$, was reasonably 
uniform.  If this condition is not satisfied, either the cost function must 
be modified or a change of variable applied to replace the attributes with
ones for which this condition is met.  It may also be necessary to exclude 
outliers.

\section{Implementation of the Bayesian Block Algorithm}

Once a cost function such as Equation (\ref{eq:total_cost}) is available, it
must be applied in a systematic fashion to an appropriate set of possible 
partitions to identify the optimum one.  The most direct way to accomplish 
this would be to evaluate the cost functions for every possible configuration 
of adjacent blocks.  Unfortunately, this approach will scale exponentially in
N, and become impractical for SOMs of any significant size.  One 
alternative is a split-and-merge algorithm similar to ones used for image 
segmentation.  While these algorithms can be somewhat ad hoc, and have the 
potential to be greedy, they offer the advantages of speed and simplicity.

We used a conventional quadtree splitting procedure, in which the decision to 
subdivide a region was based on the relative cost function of the region as a 
whole compared with four subregions.  Many different merge algorithms are
possible, but for reasons of clarity, we used the procedure described below.

\begin{enumerate}
\item{Order the regions sequentially, first by column and then by row, 
beginning in the upper left.}
\item{Examine successive regions, $R_i$, beginning with $R_1$.}
\item{For each region, examine successive neighbors, $R_{j>i}$, and compare
the cost functions, $C_{separate} = C(R_i) + C(R_j)$ and 
$C_{joined} = C(R_i \cup R_j)$ to see if the regions can be merged.}
\item{Renumber the merged regions, increment $i$, and repeat Steps 2 and 3
until no more changes occur.}
\end{enumerate}

In its current form, our implementation assumes that most of the PEs will 
not be empty, and will contain a sufficient number of samples to obtain a
meaningful measure of the standard deviation in that PE.  This restriction
may limit its ability to identify fine structure, though this issue could
be addressed through the use of a suitable magnification scheme 
[{\em Bauer et al.}, 1996].  
As noted above, this algorithm has the potential to be greedy, but this is 
unlikely to be problem for SOMs of moderate size, for which the range of
possible partitionings is severely constrained by the number of PEs and the 
restrictions imposed by the SOM algorithm itself.  

\section{Results}

We used the well-known Iris Data Set 
[{\em Blake and Men}, 1998] 
to demonstrate the usage of our Bayesian Block-based partitioning scheme and 
to compare it with more conventional schemes.  This data set, consisting of 
150 samples with 4 attributes (sepal length, sepal width, petal length, and 
petal width), divided into 3 classes of 50 samples each (iris setosa, 
versicolour, and virginica), has several qualities that make it attractive 
for the evaluation of unsupervised classification schemes.  It is 
well-understood.  It is small enough so that results can be examined in 
detail, but large enough to provide good statistics.  One of the classes 
(setosa) is cleanly separated from the others while the remaining two 
classes (versicolour and virginica) share a complicated and ill-defined 
boundary.

SOMs were generated using the Neuralware package, which is discussed at 
length by {\em Mer\'{e}nyi} [1998].  This package can use a variety of different 
neighborhood schemes and implements the 'conscience' algorithm proposed by 
{\em DeSieno} [1988] to prevent any particular PE from representing too much 
of the input data.  Classifications were performed using a 5x5 array of PEs.  
Neighborhoods were rectangular, and decreased in size from 3x3 to 1x1 during 
training.  Multiple classifications were performed to generate different 
maps.  These maps were partitioned using different values for the range and 
standard deviation parameters in Equation (\ref{eq:cost_1}) to evaluate the 
sensitivity of the algorithm to these parameters.  These partitionings were 
also compared with the best possible partitioning and the results of a 
conventional threshold-based scheme.

\begin{figure}[ht]
\centerline{\includegraphics[width=10cm,clip=]{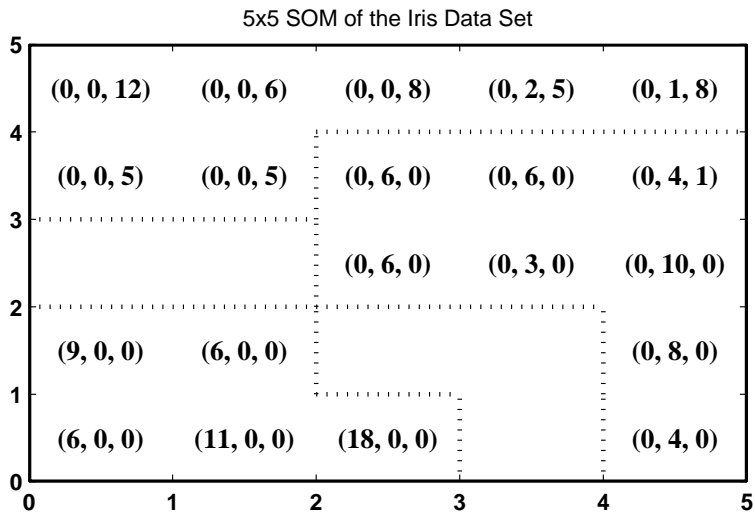}}
\caption{A representative classification of the IRIS Data Set, showing
the map produced by the SOM, the populations for each PE, and the best
possible partitioning determined using knowledge of the actual 
classifications.}
\label{fig:Figure_01}
\end{figure}

Figure \ref{fig:Figure_01} shows the map produced by a 5x5 SOM.  Populations 
for each of the three different classes are listed in parentheses for each 
PE.  The boundaries of the best possible partitioning, determined using a 
priori knowledge of the actual classifications, are shown by dotted lines.  
This partitioning classifies 146 of the 150 samples correctly, for a success 
rate of 97.33\% with a $\kappa$ of 0.96.  Different maps produced by the
multiple classification runs described above all had comparable performance.

\begin{figure}[ht]
\centerline{\includegraphics[width=14cm,clip=]{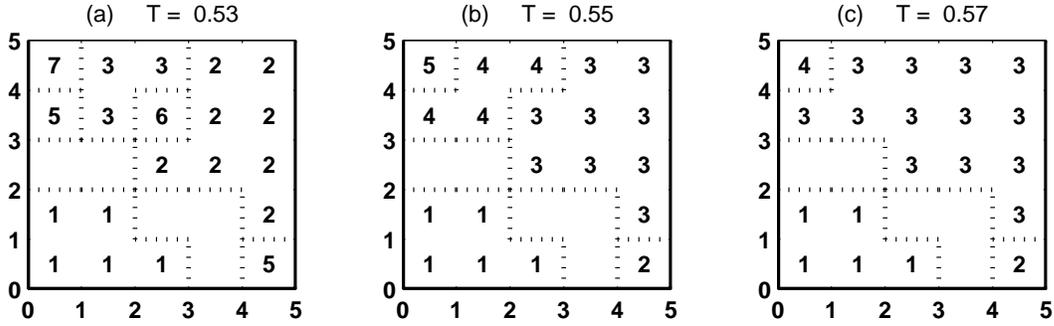}}
\caption{Threshold-based partitionings of the SOM in Figure 
\ref{fig:Figure_01} for three different values of the threshold, T: 
(a) T=0.53, (b) T=0.55, (c) T=0.57}
\label{fig:Figure_02}
\end{figure}

Figure \ref{fig:Figure_02} shows partitionings produced using a simple 
threshold-based approach similar to the u-matrix scheme of
{\em Ultsch and Simmons} [1992].
Boundary strengths were determined using the Euclidean difference between 
the mean values of the attributes of adjacent PEs (as opposed to the weight 
vectors of the PEs).  The three panels show partitionings produced using 
different thresholds for class boundaries: 0.53, 0.55, and 0.57.  All three 
partitionings overclassified the data, producing 7, 5, and 4 classes with
success rates of 64\%, 80\%, and 48\%, respectively.  The threshold of 0.55 
gave the best partitioning, classifying 120 of the samples correctly with a 
success rate of 80\% and a $\kappa$ of 0.7244.  If one knew to combine 
classes 2 and 4 and classes 3 and 5, these values would improve to 
90.7\% and 0.86, but in general, this type of additional knowledge will not 
be available.  It should also be noted that this partitioning scheme is 
extremely sensitive to the correct choice of a threshold.  A variation of 
less than 5\% resulted in a severe degradation in performance.

\begin{figure}[ht]
\centerline{\includegraphics[width=16cm,clip=]{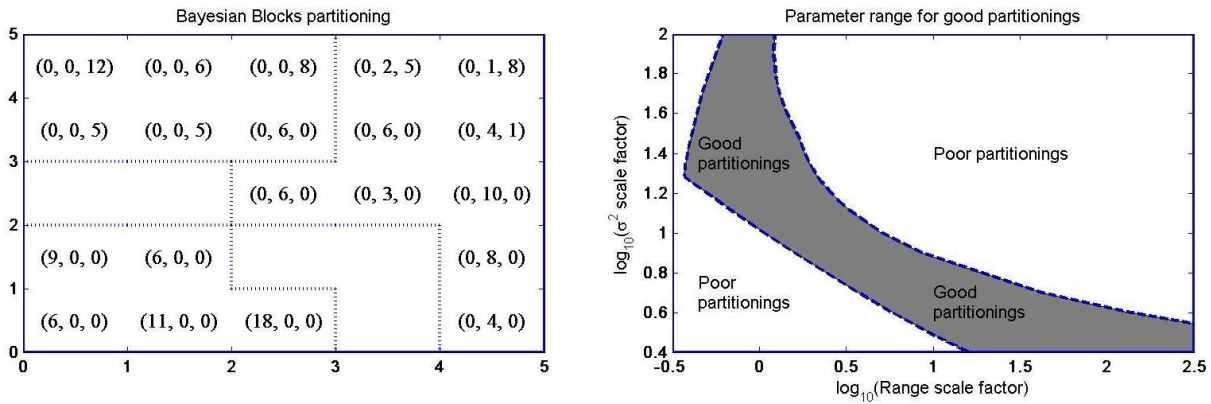}}
\caption{Bayesian Block partitioning of the SOM in Figure 
\ref{fig:Figure_01}.  Left panel: the best partitioning produced by the
Bayesian Blocks scheme.  Right panel: region of R and $\sigma$ parameters 
that reproduced this partitioning.}
\label{fig:Figure_03}
\end{figure}

Figure \ref{fig:Figure_03} shows the results of the Bayesian Block-based 
partitioning scheme described above.  The left panel of this figure shows 
the best partitioning produced by this scheme.  For this particular run, 
the range values for the four attributes were arbitrarily chosen to be twice 
the maximum observed values of 7.9, 4.4, 6.9, and 2.5 respectively, while 
the values of $\vec{\sigma}$ for each PE were taken to be 
$12 s_i \sqrt{N_{J}}$.  This partitioning identified 3 classes and assigned 
130 samples correctly for a success rate of 86.7\% and a $\kappa$ of 0.80.  
While this may not quite be as good as the best possible partitioning, it is 
significantly better than the results of the threshold-based scheme.  It is 
also significantly more robust.  The right panel of the figure shows the 
region of range and $\sigma$ parameters that reproduced this partitioning.
Note that this is quite broad.  Unlike the threshold-based scheme, which was 
sensitive to changes of as little as 5\% in the value of the threshold, the 
Bayesian Block scheme can tolerate variations in the range and $\sigma$ 
parameters of up to an order of magnitude with no effect on the results.

\section{Conclusion}

We have described a partitioning scheme for SOMs based on the Bayesian Blocks 
algorithm of {\em Scargle} [1998] and applied it to a small sample data set 
to illustrate its operation.  While a comprehensive series of performance 
evaluations on wide range of data sets is outside the scope of this paper, 
the example described here is sufficient to illustrate its usage, and 
demonstrate that it is robust.  In the future, we plan to evaluate this 
scheme with a wider range of actual and simulated data to characterize its 
performance.  As part of this process, we will explore the use of bootstrap 
methods to estimate the error and bias of the overall classification scheme.  
Our goal is to evaluate the sensitivity limitations of this partitioning
scheme, establish how it behaves when populations of different classes are 
significantly different in size, and determine how it might be affected by 
changes in the SOM magnification factor.  We also plan to investigate 
alternatives to the split-and-merge algorithm described here.  In particular, 
the higher dimensional algorithm of {\em Jackson and Scargle} [2008] can be 
applied to yield an optimal two-dimensional or higher partitioning over the 
set of all possible partitions.

This scheme requires a suitable formulation of the cost function.  As noted
above, there are several ways that this could be improved.  The estimate of 
the variation of attribute values, $\vec{\sigma}_i$, within each PE is 
somewhat ad hoc and could be replaced by a more formal procedure.  The 
simplifying assumptions regarding the prior for the attributes of a block
used in the derivation of Equation (\ref{eq:P_set_xi}) may be unduly 
restrictive and could be relaxed to include other distributions.  Finally, 
it would be desirable to devise ways to identify outliers and deal with 
them in a systematic fashion.

The Bayesian Block-based approach to partitioning falls conceptually 
somewhere between the conventional threshold-based, partitive, and 
hierarchical approaches.  As such, it partakes of many of the advantages of 
each.  Like the threshold-based and hierarchical approaches, it does not 
depend on any assumptions regarding the number and distribution of clusters 
in the sample space.  Like many partitive approaches, it is robust, stable, 
and comparatively insensitive to the choice of the parameters it uses.  Also, 
because the partitioning is based on statistics of the entire data set, it 
should also be comparatively insensitive to local variations in the data.  
These qualities should make the Bayesian Block-based approach an attractive 
alternative to conventional SOM partitioning schemes.

Sample code with documentation and a copy of the example discussed in this 
paper is available at \newline
http://astrophysics.arc.nasa.gov/\verb+~+pgazis/Bayes\_SOM/Bayes\_SOM.htm.

\section{References}

Alahakoon, L. D., S. K. Halgamuge, and B. Srinvasan,
Dynamic Self-Organizing Maps with Controlled Growth for Knowledge Discovery,
IEEE Transactions on Neural Networks 11(3), 601-614, 2000.

Bauer, H., and T. Villman,
Growing a Hypercubical Output Space in a Self-Organizing Feature Map,
IEEE Transactions on Neural Networks 3(4), 570-579, 1996.

Bauer, H., R. Der, and M. Hermann,
Controlling the Magnification Factor of Self-Organizing Maps,
in {\em Neural Computation}, vol 8, no 4, pp 757-771, 1996.

Bauer, H., M. Hermann, and T. Villmann,
Neural Maps and Topographic Vector Quantization,
Neural Networks, Vol 12, pp 659-676, 1999.

Blake, C. L. and C. J. Men, 
UCI Repository of Machine Learning Databases \newline
[http://www.ics.uci.edu/\verb+~+mlearn/MLRepository.html],
Irvine, CA: University of California, 
Department of Information and Computer Science,
1998.

DeSieno, D., Adding a Conscience to Competitive Learning,
in {\em Proc. Int. Conf. on Neural Networks, Vol 1}, 
pp 117-124, IEEE Press, NY, 1988.

Hsu, A. L. and S. K. Halgamuge,
Enhanced Topology Preservation of Dynamic Self-Organizing Maps for Data
Visualization,
in {\em Proceedings of the Joint 9th IFSA World Congress and 20th NAFIPS 
International Conference 2001}, pp 1786-1791, Vancouver, Canada, 2001.

Jackson, B. W. and J. D. Scargle,
"Optimal Partitions of Higher Dimensional Data," 
in preparation, 2008. 

Kohonen, T., {\em Self-Organizing Maps},
Springer-Verlag, Berlin, Heidelberg, New York, 1997.

Kohonen, T., {\em Self-Organization and Associative Memory},
Springer-Verlag, Berlin, 1984.

Mer\'{e}nyi, E., Self-Organized ANNs for Planetary Surface Composition
Research, 
in {\em Proc. European Symposium on Artificial Neural Networks,
ESANN98}, pp 197-202, Bruges, Belgium, 1998.

Merkl, D. and A. Rauber, A.,
Cluster connections - a visualization technique to reveal cluster boundaries 
in self-organizing maps.
in {\em Proc 9th Italian Workshop on Neural Nets (WIRN97)}, 
Vietri sul Mare, Italy, 1997.

Ritter, H., T. Martinez, K. Schulten,
{\em Neural Computation and Self-Organized Maps},
Addison-Wesley, Reading, Mass., 1992.

Scargle, J. D., 1998, Studies in Astronomical Time Series Analysis. V. 
Bayesian Blocks, A New Method to Analyze Structure in Photon Counting Data", 
{\em Ap. J.}, {\em 504}, p.405-418, 
Paper V. http://xxx.lanl.gov/abs/astro-ph/9711233. 1998.

Scargle, J. D., Bayesian Blocks in Two or More Dimensions: Image 
Segmentation and Cluster Analysis," 
Contribution to Workshop on Bayesian Inference and Maximum Entropy Methods 
in Science and Engineering (MAXENT 2001), 
held at Johns Hopkins University, Baltimore, MD USA on August 4-9, 2001.

Scargle, J. D., J. Norris, B. Jackson,
Studies in Astronomical Time Series Analysis. VI.
Optimal Segmentation: Blocks, Histograms and Triggers,
2008.

Ta\c{s}demir K., and E. Mer\'{e}nyi,
Considering Topology in the Clustering of Self-Organizing Maps,
in {\em Proc. 5th Workshop On Self-Organizing Maps (WSOM 2005)}, 
Paris, France, pp 439-446, 2005. 

Ultsch, A. and H. P. Siemon, Kohonen Neural Networks for Exploratory
Data Analysis, in {\em Proc. Conf. Soc. for Information and
Classification}, Dortmund, 1992.  

Vesanto, J, and E. Alhoniemi,
Clustering of the Self-Organizing Map,
IEEE Transactions on Neural Networks, Vol 11, 586-200, 2000.

Villmann, T., M. Hermann, R. Der, and M. Martinetz,
Topology Preservation in Self-Organizing Feature Maps: Exact Definition and
Measurement,
IEEE Transactions on Neural Networks, Vol 8, No. 2, 1997.

Villmann, T. and E. Mer\'{e}nyi, Extensions and Modifications of the 
Kohonen-SOM and Applications in Remote Sensing Image Analysis,
in {\em Self-Organizing Maps, Recent Advances and Applications}
U. Seiffet and L. C. Jain, eds, Springer Verlag, Berlin, 
121-145, 2001.

\end{document}